\documentclass[a4paper,preprint]{revtex4-1}
\pdfoutput=1
\usepackage{graphicx}
\usepackage{amsmath}
\usepackage{fancyhdr}
\usepackage{setspace}
\usepackage{amssymb}
\usepackage[margin=1in]{geometry}
\usepackage{color}
\usepackage{natbib}

\begin{document}
\title{Entropy production rate as a criterion for inconsistency decision theory}
\author{Purushottam D. Dixit}
\affiliation{Department of Systems Biology, Columbia University\\ New York, NY}
\thanks{Correponding author. Email: dixitpd@gmail.com}

\setstretch{1.5}

\begin{abstract}
Individual and group decisions are complex, often involving choosing an apt alternative from a multitude of options. Evaluating pairwise comparisons breaks down such complex decision problems into tractable ones. Pairwise comparison matrices (PCMs) are regularly used to solve multiple-criteria decision-making (MCDM) problems, for example, using Saaty's analytic hierarchy process (AHP) framework. However, there are  two significant drawbacks of using PCMs. First, humans evaluate PCMs in an inconsistent manner.  Second, not all entries of a large PCM can be reliably filled by human decision makers.  We address these two issues by first establishing a novel connection between PCMs and time-irreversible Markov processes. Specifically, we show that every PCM induces a family of dissipative maximum path entropy random walks (MERW) over the set of alternatives. We show that only `consistent' PCMs correspond to detailed balanced MERWs. We identify the non-equilibrium entropy production in the induced MERWs as a metric of inconsistency of the underlying PCMs.  Notably, the entropy production satisfies all of the recently laid out criteria for reasonable consistency indices. We also propose an approach to use incompletely filled PCMs in AHP.  Potential future avenues are discussed as well.

keywords: analytic hierarchy process, markov chains, maximum entropy
\end{abstract}

\maketitle

\section{Introduction}

Individuals and organizations regularly have to choose an `optimal' alternative from a large number of available options. Often, the individual alternatives have multiple attributes (for example, cost, quality, and durability) which makes the decision complex. On the one hand, if only single attributes are considered individual alternatives can be ranked on a one dimensional absolute preference scale. On the other hand, no such scale may exist when all  attributes are considered.  Nevertheless, it has been argued that it is possible for human agents to robustly compare pairs of alternatives in a high dimensional attribute space~\citep{saaty2012models}.  Pairwise comparison matrices (PCMs) were first introduced in a nascent form in psychophysics by Fechner in the 1860~\citep{fechner2012elemente} and later rigorously defined by Thurstone in the 1920s~\cite{thurstone1927law}. PCMs allow agents to simplify complex decision making problems by breaking them up into smaller tractable ones. Starting from the 1970s, Saaty devised a framework to approximate the absolute preference scale from PCMs using his analytic hierarchy process (AHP)~\cite{saaty2004decision,saaty2012models} and analytic network process (ANP)~\cite{saaty2004decision}.  

Mathematically, PCMs are organized as follows. Consider that an agent has to choose from $n > 1$ alternatives denoted by $\{ a\}$.  The entry $W_{ab} \in \mathbb{R}_{>0}$ in a PCM $W$ denotes the preference of an agent for alternative $a$ over alternative $b$. The cross-diagonal entries of the PCM are reciprocals of each other;  $W_{ba} = 1/W_{ab}$.

While PCMs simplify large complex problems, two drawbacks have been identified. First, agents may decide between pairs of alternative using a combination of quantitative analysis and qualitative intuition. As a result, not all pairwise comparisons within a matrix may be `consistent' with each other~\cite{saaty2012models}. For example, if an agent prefers `$a$' over  `$b$' by a factor of 2 and `$b$' over `$c$' by a factor of 2, in real PCMs, it is not guaranteed that the same agent will also prefer `$a$' over `$c$' by a factor of $W_{ab}W_{bc} = 4$.  For a consistent PCM, we have for any path over the alternatives $a_1 \rightarrow a_2 \rightarrow \dots a_k$
\begin{eqnarray}
\log W_{a_{1}a_{k}} = \sum_{i=1}^{k-1} \log W_{a_{i}a_{i+1}}. \label{eq:consist}
\end{eqnarray}
Consequently, the individual entries of a consistent PCMs can be expressed as~\citep{saaty2012models}
\begin{eqnarray}
\log W_{ab} =\log f_a - \log f_b
\end{eqnarray}
for some absolute preference scale $\bar f > 0$~\cite{saaty2012models}. In other words,  individual pairwise comparisons  in a consistent PCM can be represented by a `state function' $\log \bar f$. Notably, the absolute preference scale is proportional to the right Perron-Frobenius eigenvector $\bar \nu$ if the PCM is consistent ($\nu_a \propto f_a$).  Based on the relationship between the Perron-Frobenius eigenvector and the absolute preference scale for consistent PCMs, Saaty in his AHP argued that the same eigenvector also approximates the absolute preference scale for inconsistent PCMs~\citep{saaty2004decision,saaty2012models}.  The AHP approach is now regularly used to infer absolute preference scales over alternatives in a broad range of areas such as environmental sciences~\citep{ramanathan2001note}, organizational studies~\citep{nydick1992using}, and public health~\citep{liberatore2008analytic}. %

In addition to their use in AHP, PCMs also allow agents to identify the sources of departure from consistency in individual and group decision making~\citep{brunelli2013inconsistency}.   Over the last three decades, several  indices have been developed to quantify consistency of PCMs~\citep{brunelli2013inconsistency}.  For example, a popular index by Saaty~\citep{saaty2012models} quantifies the Perron-Frobenius eigenvalue $\eta$ of the PCM. Saaty showed that  $\eta \ge n$ for any PCMs with equality holding {\it iff} the PCM is consistent. Based on this observation, he defined the consistency index ${\rm CI}$:
\begin{eqnarray}
{\rm CI} = \frac{\eta-n}{n-1}. \label{eq:CI}
\end{eqnarray}
Other examples of consistency indices include the Harmonic consistency index~\cite{stein2007harmonic} and Geometric consistency index~\citep{aguaron2003geometric}.  Recently Brunelli et al.~\cite{brunelli2015axiomatic,brunelli2017studying} laid out a set of requirements for reasonable quantifiers of departures from consistency of PCMs.

Second, several entries in a large PCM ($n \gg 1$) may be missing. The reasons are several fold. The total number of pairwise comparisons ${n \choose 2}$ increases as the square of the total number of alternatives $n$. Indeed, psychological studies have shown that human agents are not able to reliable estimate multiple pairwise comparisons at the same time because of information overload or simply because they get bored and/or inattentive (see~\citep{carmone1997monte} references). Moreover, not all comparisons may be realistically available (for example when ranking sports teams or players with non-overlapping stints~\citep{csato2013ranking,bozoki2016application}).  A number of proposals fill up incomplete PCMs using the available entries have been suggested~\citep{harker1987incomplete,harker1987alternative,koczkodaj1999managing,fedrizzi2007incomplete,bozoki2010optimal}. A popular proposal by Harker~\citep{harker1987incomplete} is as follows. We define the adjacency graph matrix $A$ of an incompletely filled PCM $W$.  We have 
\begin{eqnarray}
A_{ab} = 1~{\rm if}~W_{ab} > 0~{\rm and}~A_{ab}=0~{\rm if}~W_{ab} = 0.
\end{eqnarray}
We assume that the adjacency graph is connected. For any missing entry, say $W_{ab}$, we first enumerate all possible elementary paths $\{ \Gamma \}$ on $A$ between $a$ and $b$. Next, we approximate the comparison along each path $\Gamma \equiv a\rightarrow a_1 \rightarrow a_2  \rightarrow \dots \rightarrow a_k \rightarrow  b$ {\it as if} the known entries in the PCM were described by a state function,
\begin{eqnarray}
\log W_{ab} (\Gamma) &=& \log W_{aa_{1}} +  \left (\sum_{i=1}^{k-1} \log  W_{a_{i}{a_{i+1}}} \right ) \nonumber \\ &+& \log W_{a_{k}b}.
\end{eqnarray}
The logarithm of the missing entry $\log W_{ab}$ is then approximated as an arithmatic mean of $\log W_{ab}(\Gamma)$ over all possible elementary paths between $a$ and $b$. All previously missing entries defined this way automatically satisfy $W_{ab}W_{ba} = 1$. Notably, most subsequent analyses of PCMs assume that they are positive ($W_{ab} > 0~\forall~a$ and $b$). Thus, it is not clear how any particular filling up proposal may bias the sbubsequent analyses of PCMs. 

In this work we address the following problem: is there a way to analyze PCMs without relying on specific proposals to fill them up? We provide a physics-based answer.  First, we establish a novel connection between PCMs and and time-irreversible statistical physics.  Specifically, we show that every PCM (incompletely filled or otherwise) induces a family of Markovian random walks over the alternatives. The random walks are maximum path entropy random walks constrained to reproduce a  `flux' $J \propto \langle \log W_{ab} \rangle$~\cite{burda2009localization,frank2014information,dixit2015stationary}. This connection allows us to bring insights from recent work in stochastic thermodynamics~\citep{seifert2008stochastic} to study of PCMs. Notably, we show that the entropy production rate in the induced random walks is intricately related to the consistency of the underlying PCM. Quantification of entropy production does not require filling up of the PCM as long as the adjacency graph $A$ of the PCM is connected. Moreover,  the entropy production can be decomposed as either a sum over (a) alternatives or (b) pairwise comparisons which allows us to directly identify the alternatives or the comparisons that are inconsistent with the rest of the PCM. We  provide physics-based explanations for previously laid out conditions for reasonable inconsistency indices~\citep{brunelli2015axiomatic,brunelli2017studying}. We also show that the absolute preference scale can be extracted from an incompletely filled but otherwise consistent PCM by correcting for the effect of the adjacency graph of the incomplete PCM. This allows us to generalize Saaty's AHP for incompletely filled PCMs. 

We numerically compare our consistency index with previously developed ones. We illustrate our development by systematically examining the effect of filling up incompletely filled matrices on the evaluation of consistency and the AHP. We show that filling up a PCM introduces systematic biases in evaluation of consistency. Importantly, we believe that the connections established in this work between two previously unrelated fields of scientific inquiry will allow a greater understanding of consistency of pairwise comparison matrices in the future.

\section{Results}

\subsection{PCM-induced random walk}

Consider an incompletely filled PCM $W$. We assume that the unquerried entries in $W$ are set to zero and  $W_{ab} > 0 \Rightarrow W_{ab}W_{ba} = 1$. We also assume that the  adjacency graph matrix $A$ of $W$ is connected. We define a microscopic `flux' $j_{ab} = \log W_{ab}$ between vertices $a$ and $b$ of $A$ that have an edge between them. Note that the microscopic flux is antisymmetric; $j_{ab} + j_{ba} = 0.$ 

We construct a discrete time Markov process with transition probabilities $\{ k_{ab} \}$ and a stationary distribution $\{ p_a \}$ on $A$ that is consistent with a given ensemble average flux $J = \langle j_{ab} \rangle$ per unit time. The ensemble average flux is given by
\begin{eqnarray}
\langle j_{ab} \rangle = \sum_{a,b} p_a k_{ab} j_{ab}.
\end{eqnarray}

There are infinitely many Markov processes consistent with a single path ensemble average. We seek the one with the maximum path entropy. The path entropy is given by~\cite{dixit2014inferring,dixit2015inferring,dixit2015stationary}
\begin{eqnarray}
S = -\sum_{a,b} p_a k_{ab} \log k_{ab}.
\end{eqnarray}
Maximization of $S$ is a constrained  problem~\cite{dixit2014inferring,dixit2015inferring,dixit2015stationary} because $\{ p_a \}$ and $\{ k_{ab} \}$ are dependent of each other,
\begin{eqnarray}
\sum_b p_a k_{ab} &=& p_a,~\sum_{a,b} p_a k_{ab} =  1,~\sum_a p_a k_{ab} = p_b
 \label{eq:c1}
\end{eqnarray}
and 
\begin{eqnarray}
\sum_{a,b} p_a k_{ab} j_{ab} = \langle j_{ab} \rangle = J \label{eq:c2}
\end{eqnarray}
Eqs.~\ref{eq:c1} represent the constraint of probability conservation and normalization and $\{ p_a \}$ as the stationary distribution respectively. Eq.~\ref{eq:c2} represents the imposed path ensemble constraint of the flux $J$. We solve the constrained problem using the method of Lagrange Multipliers. We write the unconstrained Caliber~\cite{dixitMaxCal}
\begin{eqnarray}
C = S &+& \sum_a l_a \left ( \sum_{b} p_a k_{ab} - p_a\right ) + \sum_b m_b \left ( \sum_{a} p_a k_{ab} - p_b\right ) +  \delta \left (\sum_{ab} p_{a} k_{ab} - 1 \right ) \nonumber \\ &+& \gamma \left ( \sum_{a,b} p_a k_{ab} j_{ab} - J \right )  \label{eq:caliber}
\end{eqnarray}
In Eq.~\ref{eq:caliber}, $\{ l_a \}$, $\{ m_b \}$, and $\delta$ are the Lagrange multipliers that impose the constraints in Eq.~\ref{eq:c1}. $\gamma$ is the Lagrange multiplier that imposes the dynamical flux constraint. The transition probabilities of maximum path entropy random walks (MERW) are given by~\cite{dixit2015stationary}
\begin{eqnarray}
k_{ab}(\gamma) = \frac{\nu_b(\gamma)}{\eta(\gamma) \nu_a(\gamma)} { W}_{ab}^{\gamma} \label{eq:merw}
\end{eqnarray}
where $\eta(\gamma)$ is the largest eigenvalue of the modified PCM ${W}^{\gamma}$, and $\bar \nu(\gamma)$ is the corresponding right eigenvector. From here onwards, we recognize $W^{\gamma}$ as the element-wise exponentiation and not the matrix exponentiation.  According to the Perron-Frobenius theorem, $\bar \nu(\gamma)$ has positive entries and $\eta(\gamma) > 0$. 
Finally, the stationary distribution $\{ p_a(\gamma) \}$ is given by the outer product~\cite{dixit2015stationary}
\begin{eqnarray}
p_a(\gamma) = \nu_a(\gamma) \mu_a(\gamma)
\end{eqnarray}
where $\mu_a(\gamma)$ is the left Perron-Frobenius eigenvector of ${W}^{\gamma}$. We call the $\gamma$-dependent family of Markov processes described by Eq.~\ref{eq:merw} the maximum path entropy random walks induced by the PCM.  Notably, both the left and the right Perron-Frobenius eigenvectors are used as an approximate absolute preference scale in Saaty's  AHP~\citep{saaty2004decision,saaty2012models}. 

We note that the Markov process  at $\gamma = 1$ corresponds to the original PCM $W$. From here onwards, unless specified otherwise we will assume that $\gamma = 1$. We omit the dependence on $\gamma$ for brevity.

\subsection{Entropy production as a metric of inconsistency}

The entropy production rate of a Markov process quantifies the degree of irreversibility in it; the entropy production rate is zero {\it iff} the process is time-symmetric and satisfies detailed balance $p_a k_{ab} = p_b k_{ba}~\forall~a$ and $b$~\cite{seifert2008stochastic}. We have~\cite{dixit2015stationary}
\begin{eqnarray}
\dot s &=& \sum_{a,b} p_a k_{ab} \log \frac{k_{ab}}{k_{ba}} \\
&=& \sum_{a,b} p_a k_{ab} \left ( \log \frac{\nu_b^2}{\nu_a^2}  + 2 \log  W_{ab} \right ) \nonumber \\ &=&  \frac{2}{\eta}\sum\mu_a \nu_b W_{ab} \log W_{ab} = 2\langle j_{ab} \rangle \ge 0.\label{eq:sdot}
\end{eqnarray}
Note that the entropy production rate in Eq.~\ref{eq:sdot} is evaluated at $\gamma = 1$.  For an arbitrary $\gamma$, we have 
\begin{eqnarray}
\dot s(\gamma) = 2\gamma \langle j_{ab}\rangle(\gamma)
\end{eqnarray}
where $\langle j_{ab} \rangle(\gamma)$ is the flux in the induced MERW when the Lagrange multiplier is set at $\gamma$.

The entropy production rate $\dot s$ of the induced MERW defined in Eq.~\ref{eq:sdot} serves as a physics-based quantifer of the inconsistency of the underlying PCM. We prove that $\dot s  = 0$ {\it iff} the underlying PCM is consistent. First, consider $\dot s = 0$. We show that all non-zero entries of $W$ are given by $W_{ab} = f_a/f_b$ for some absolute scale $\bar f > 0$. Detailed balance implies 
\begin{eqnarray}
\mu_a \nu_a \frac{\nu_b}{\eta \nu_a} { W}_{ab} &=&\mu_b \nu_b \frac{\nu_a}{\eta \nu_b} { W}_{ba} \\
\Rightarrow \frac{W_{ab}}{W_{ba}} &=& W_{ab}^2 =   \frac{\mu_b}{\nu_b} \times \frac{\nu_a}{\mu_a}  \\
\Rightarrow W_{ab} &=& \frac{f_a}{f_b}
\end{eqnarray}
where $f_a = (\nu_a/\mu_a)$. 

Next,  consider a PCM $W$ whose  non-zero entries are given by $W_{ab} = f_a/f_b$ where $\bar f > 0$ is a vector of positive elements. We evaluate the entropy production rate of the induced MERW. First, we derive the transition probabilities $k_{ab}$. We write
\begin{eqnarray}
W  &=& {\rm Diag}(\bar f ) A {\rm Diag}(1/ \bar f) \\
\Rightarrow W{\rm Diag}(\bar f ) &=& {\rm Diag}(\bar f)A \label{eq:WA}
\end{eqnarray}
In Eq.~\ref{eq:WA}, Diag$(\bar x)$ is the diagonal matrix with entries from the vector $\bar x$ and $A$ is the adjacency matrix of $W$. Let $\eta$ be the Perron-Frobenius eigenvalue of $A$ and $\bar \nu$ be the corresponding Perron-Frobenius eigenvector. We have
\begin{eqnarray}
A\bar \nu &=& \eta \bar \nu \\
\Rightarrow W{\rm Diag}(\bar f) \bar \nu &=& {\rm Diag}(\bar f)A \bar \nu \nonumber \\ &=& \eta  {\rm Diag}(\bar f)\bar \nu. \label{eq:incomplete_ahp0}
\end{eqnarray}  
In other words, the Perron-Frobenius eigenvalue of $W$ is $\eta$ and the corresponding eigenvector is $\bar g$ where $g_a = \nu_a f_a$. The transition probabilities of the MERW  are given by (see Eq.~\ref{eq:merw})
\begin{eqnarray}
k_{ab} = \frac{\nu_b f_b }{\eta \nu_a f_a} \times f_a /f_b A_{ab} = \frac{\nu_b}{\eta \nu_a}A_{ab} \label{eq:merw_special}
\end{eqnarray}
where $A_{ab} = 1$ if $a$ and $b$ are connected by an edge (if the comparison between $a$ and $b$ is available) and zero otherwise. Since $A$ is symmetric ($A_{ab} = A_{ba}~\forall~a$ and $b$), the Markov process described by Eq.~\ref{eq:merw_special} satisfies detailed balance and has a zero entropy production rate~\cite{burda2009localization,frank2014information,dixit2015stationary}. In other words, the induce MERW is detailed balanced {\it iff} the underlying PCM is consistent.  Notably,  Brunelli's first requirement for any metric that measures inconsistency is the ability to uniquely identify consistent PCMs~\cite{brunelli2015axiomatic,brunelli2017studying}. As shown here, the entropy production rate satisfies this requirement.

Why does $\dot s$ quantify consistency? Let us consider the detailed balanced Markov process induced by a consistent PCM. As we showed above, the MERW induced by a consistent PCM is detailed balanced ($p_a k_{ab} = p_b k_{ba}~\forall~a$ and $b$). An illustrative analogy is to imagine that the MERW describes a system at thermodynamic equilibrium with a surrounding bath. Let us assume that the stationary distribution of the MERW is given by $p_a \propto e^{-\beta E_a}$ where $E_a$ is the `energy' of the alternative $a$. Let us consider a path over the alternatives $\Gamma = a_1 \rightarrow a_2 \rightarrow \dots \rightarrow a_m$ and the corresponding time reversed path $\Gamma' = a_m \rightarrow a_{m-1} \rightarrow \dots \rightarrow a_1$. 

The log ratio of the forward and the reverse path probabilities is related to the total heat exchanged during the trajectory. We have,
\begin{eqnarray}
\log  \frac{p(\Gamma|a_1)}{p(\Gamma'|a_m)} &=& \log  \frac{k_{a_{1}a_{2}} \times \dots \times k_{a_{m-1}a_{m}}}{k_{a_{m}a_{m-1}} \times \dots \times k_{a_{2}a_{1}}} \\
&=& \log \left ( \frac{p_{a_{2}}}{p_{a_{1}}} \times \dots \times \frac{ p_{a_{m}} }{ p_{a_{m-1}} }  \right ) \\ \nonumber \\ &=& \Delta Q = -\beta \left ( E_m - E_1 \right ) \label{eq:dissipate}
\end{eqnarray}
Notably, the heat exchange is independent of the path only for detailed balance processes. Specifically, the heat exchange is zero for all loops i.e. $a_m = a_1 \Rightarrow \Delta Q = 0$. In contrast, the  heat dissipation in the MERW induced by an inconsistent PCM depends on the entire history of the trajectory. We have~\citep{dixit2015stationary}
\begin{eqnarray}
\log  \frac{p(\Gamma|a_1)}{p(\Gamma'|a_m)} &=&2\gamma \sum_{i=1}^{m-1} j_{a_{i}a_{i+1}} + 2\log \frac{ \nu_{a_{m}} }{ \nu_{a_{1}} }. \label{eq:dissipate_1}
\end{eqnarray}

To translate this observation in the language of PCMs, let us consider a Markovian random walker on a looped trajectory of the induced MERW. We imagine that the random walker exchanges `energy' with the `surrounding'. Every time step when the walker goes to an alternative that is less favored compared to the current one ($W_{a_{i}a_{i+1}} > 1$) she receives energetic renumeration $\propto \log W_{a_{i}a_{i+1}}$. However, she has to pay the same amount of energy when she goes to an alternative that is more favored. On the one hand, if the MERW is detailed balanced (if the PCM is consistent), the walker will end up with no net change in energy over {\it any} loop.  On the other hand, if the PCM is inconsistent, there will exist loops which have a net exchange of energy between the walker and the surrounding.

The induced MERWs have few other notable properties. Consider a long path $\Gamma = a_1 \rightarrow a_2 \rightarrow \dots $ of length $T \gg 1$ of the MERW for a fixed value of $\gamma$. From Eq.~\ref{eq:merw}, we write the probability $p(\Gamma)$~\cite{dixit2015stationary}
\begin{eqnarray}
p(\Gamma) \approx \frac{1}{\eta(\gamma)^{T-1}} e^{(T-1)\gamma j(\Gamma)}
\end{eqnarray}
where 
\begin{eqnarray}
j(\Gamma) = \frac{1}{T-1} \sum_{i = 1}^{T-1} j_{a_{i}a_{i+1}}
\end{eqnarray}
is the flux per unit time associated with the path $\Gamma$. We recognize $\eta(\gamma)^{T-1}$ as the partition function. Since $p(\Gamma)$ is normalized, we write (in the limit $T\gg 1$)
\begin{eqnarray}
\eta(\gamma)^{T-1} &=& \sum_\Gamma e^{(T-1)\gamma j(\Gamma)}  \\
\Rightarrow \langle j_{ab} \rangle (\gamma) &=& \frac{d}{d\gamma} \log \eta(\gamma).
\end{eqnarray}
We note that $\eta(-\gamma)$ is the Perron-Frobenius eigenvalue of $W^{-\gamma} = \left ({\rm Trans}({W})\right)^\gamma$ where Trans$(W)$ is the transpose of $W$. Since the eigenvalues of $W$ and Trans$(W)$ are the same, we conclude that $\eta(-\gamma) = \eta(\gamma)$ i.e. $\eta(\gamma)$ is an even function of $\gamma$. Consequently, it's derivative $\langle j_{ab}\rangle(\gamma)$ is an odd function of $\gamma$. Moreover,
\begin{eqnarray}
\frac{d}{d\gamma} \langle j_{ab}\rangle(\gamma) &=& \frac{1}{T-1}\frac{d^2}{d\gamma^2}  \log \sum_\Gamma e^{(T-1)\gamma j(\Gamma)} \\
&=& \langle j_{ab}^2\rangle - \langle j_{ab}\rangle^2 \ge 0.
\end{eqnarray}
Thus, $\langle j_{ab}\rangle(\gamma)$ is a monotonic function of $\gamma$. As a result, {\bf (1)} $\dot s(\gamma)$ is a monotonically increasing function of $\gamma$ for $\gamma > 0$ and since $\langle j_{ab} \rangle(\gamma)$ is an odd function of $\gamma$ {\bf (2)} $\dot s =2\gamma \langle j_{ab} \rangle(\gamma)$ is an even function of $\gamma.$ Notably, these two observations directly correspond with requirement (3) and (6) laid out by Brunelli's~\cite{brunelli2015axiomatic,brunelli2017studying}.  In appendix~\ref{ap1}, we show that $\dot s(\gamma)$ satisfies all of Brunelli's requirements.

\subsection{Using incomplete PCMs in the AHP}

Can we perform Saaty's AHP analysis on an incompletely filled PCM? As above, let us consider an incompletely filled but otherwise consistent PCM. We have 
\begin{eqnarray}
W_{ab} > 0 \Rightarrow W_{ab} = f_a/f_b .
\end{eqnarray}
Can we extract the absolute preference scale from $W$? Eq.~\ref{eq:incomplete_ahp0} shows that the Perron-Frobenius eigenvector $\bar g$ of $W$ is given by $g_a = \nu_a f_a$ where $\bar \nu$ is the Perron-Frobenius eigenvector of the adjacency matrix $A$ corresponding to $W$. Notably, the absolute preference scale can be extracted from an incompletely filled but otherwise consistent PCM not as the  reciprocal of the right Perron-Frobenius eigenvector $\bar g$,  but with a correction that accounts for the connectivity in the adjacency graph:
\begin{eqnarray}
f_a = g_a/\nu_a. \label{eq:incomplete_ahp1}
\end{eqnarray} 
Similar in spirit to the original observation of the AHP, we propose that for inconsistent and incompletely filled PCMs, the absolute preference scale $\bar f$ can be approximated using the Perron-Frobenius eigenvector $\bar g$ of the PCM $W$ and the Perron-Frobenius eigenvector $\bar \nu$ of the corresponding adjacency matrix $A$ as shown in Eq.~\ref{eq:incomplete_ahp1}.

\section{Illustrative examples}

\begin{figure}
	\includegraphics[scale=0.6]{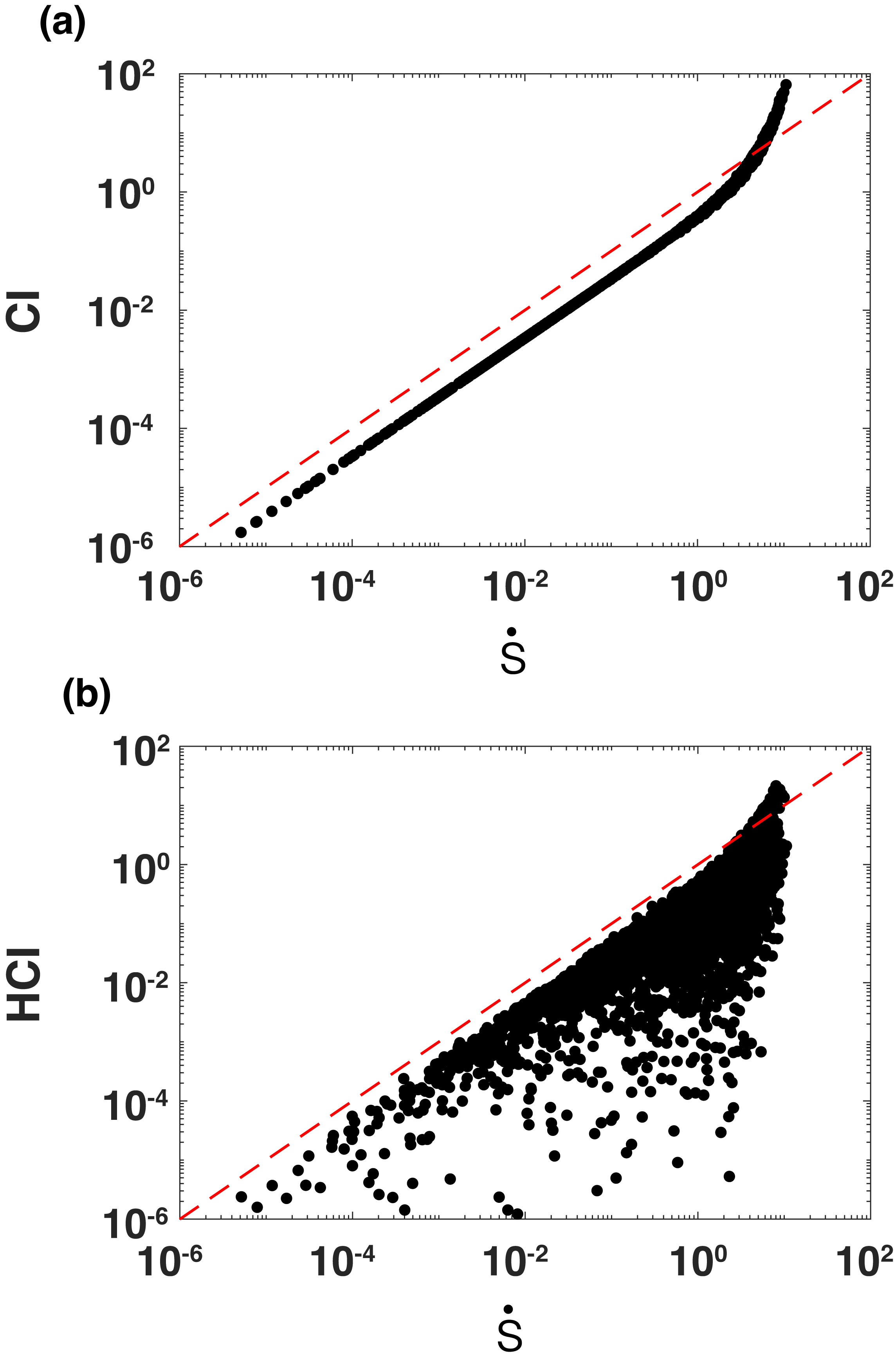}
	\caption{Panel (a) Comparison between Saaty's CI and $\dot s$ as defined in Eq.~\ref{eq:sdot} for randomly generated PCMs with varying degree of inconsistency. Panel (b) Comparison between the Harmonic consistency index HCI and $\dot s$ as defined in Eq.~\ref{eq:sdot} for randomly generated PCMs with varying degree of inconsistency. \label{fg:compare}}
\end{figure}

\subsection{Numerical comparison between $\dot s$ and other consistency indices}

We illustrate $\dot s$ as a quantifier of inconsistency by evaluating the inconsistency of multiple incompletely filled pairwise comparison matrices. Specifically, we study Saaty's CI (see Eq.~\ref{eq:CI}) and the harmonic consistency index (HCI)~\citep{stein2007harmonic}. We have already introduced Saaty's CI, here we briefly introduce the HCI and the GCI. Since consistent PCMs $W$ have rank$(W)=1$, the colums are proportional to each other. Consequently, if $t_a = \sum_a W_{ab}$, it was proven that $\sum t_a^{-1} = 1$ {\it iff} $W$ is consistent~\citep{stein2007harmonic}. The HCI quantifies deviations of the harmonic mean $HM = n/\sum t_a^{-1}$ from $n$. We have
\begin{eqnarray}
HCI = \frac{\left (HM- n \right )(n+1)}{n(n-1)}
\end{eqnarray}

We construct an ensemble of PCMs with $n=5$ of varying degree of inconsistency. First,  we construct an absolute scale $\bar f > 0$ by drawing $n=5$ uniformly distributed random numbers between [0, 1].  We construct a PCM $W$ with elements 
\begin{eqnarray}
W_{ab} = \frac{f_a}{f_b} \exp \left ( \rho_{ab} \alpha \right )~{\rm for}~a > b.
\end{eqnarray}
The parameter $\alpha$ controls the inconsistency. PCMs with $\alpha = 0$ are consistent (but incompletely filled) and the inconsistency increases as $\alpha$ increases. We choose $\alpha$ to be randomly distributed between $[0, 4].$ $\rho_{ab}$ are normally distributed random numbers with zero mean and unit standard deviation. The lower-diagonal entries of $W$ are filled to satisfy the reciprocal relationship $W_{ab}W_{ba} = 1$. 

In Fig.~\ref{fg:compare}, we compare our inconsistency index $\dot s$ with the CI and the HCI. The dashed red line shows $x=y$. Notably, $\dot s$ correlates extremely well with Saaty's CI (Pearson $r^2 = 0.98$). This may be because $\dot s$ directly depends on the Perron-Frobenius eigenvalue $\eta$ (see Eq.~\ref{eq:sdot}). $\dot s$ also correlates well with the HCI. However the correlation is lower (Pearson $r^2 = 0.71$) and there is a a large scatter.

\subsection{Inferring preference scales from incomplete matrices}

We now show how to infer the absolute preference scale using an incompletely filled PCMs using  Eq.~\ref{eq:incomplete_ahp1} as proposed above. We compare our approach with the approach by Harker~\citep{harker1987incomplete} described above.

\begin{figure}
	\includegraphics[scale=0.5]{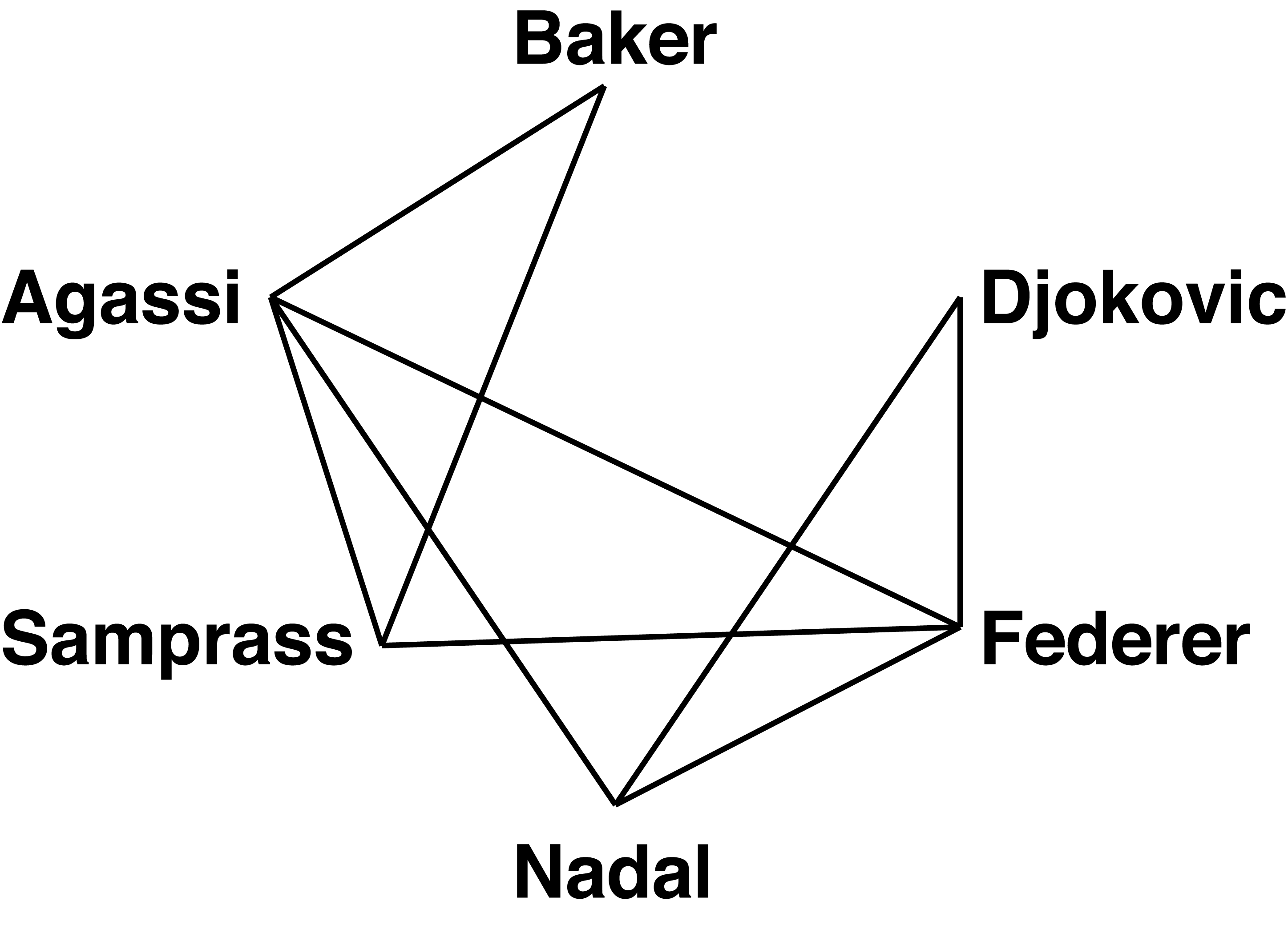}
	\caption{The graph of adjacency between Tennis players corresponding to the pairwise comparison matrix given in Table~\ref{tb:pcm}. \label{fg:graph}}
\end{figure}

Recently, Boz\'oki et al.~\citep{bozoki2016application} studied the problem of determining ranking among 25 Tennis players based on their performance against each other. Notably, not all players played with each other, for example, Agassi never played a match with Djockovic. Consequently, the PCM constructed using players' performance is inherently incomplete. Here, we carry out an analysis on 6 of the 25 players; Agassi (A), Baker (B), Djokovic (D), Federer (F), Nadal (N), and Samprass (S). The PCM is given in Table~\ref{tb:pcm} (see Boz\'oki et al.~\citep{bozoki2016application} for details). We set to zero all incompletely filled entries. The graph of connectivity among the tennis players is shown in Fig.~\ref{fg:graph}. 
\begin{table}
\begin{tabular}{c|ccccccc}
  &A & B & D & F & N & S  \\\hline
A &1 & 1.39 & 0 & 0.76 & 0.9 & 0.73\\
B & 0.72 & 1 & 0 & 0 & 0 & 0.77 \\
D & 0 & 0 & 1 & 0.95 & 0.77 & 0 \\
F &  1.32 & 0 & 1.05 & 1 & 0.52 & 1.05\\
N & 1.11 & 0 & 1.29 & 1.91 & 1& 0\\
S & 1.36 & 1.3 & 0 & 0.95 & 0 & 1
\end{tabular}
\caption{Pairwise comparison matrix among Tennis players based on their performance against each other. Entries between players who did not play any matches between them are set to zero.  \label{tb:pcm}}
\end{table}

In Table~\ref{tb:results}, we show the results of our calculations. First, we estimate the absolute preference scale using Harker's method~\citep{harker1987incomplete}. The `filled' PCM is given in Table~\ref{tb:pcmH}.  The principal eigenvector of the filled PCM $\bar f_H$ is given in column 1.  The absolute preference scale is $\mathcal L_1$ normalized.  In order to evaluate the absolute preference scale using Eq.~\ref{eq:incomplete_ahp1}, we first evaluate the right Perron-Frobenius eigenvector $\bar \nu$ (column 2). Next, we evaluate the right Perron-Frobeius eigenvector of the incompletely filled matrix in Table~\ref{tb:pcm} $\bar g$ (column 3). Finally, the $\mathcal L_1$ normalized estimate of the absolute preference scale using Eq.~\ref{eq:incomplete_ahp1} is given by $\bar f$ (column 4). Notably, $\bar f_H$ and $\bar f$ are highly correlated with each other (Pearson's $r^2 = 0.99, p=3.3\times 10^{-5}$). 
\begin{table}
\begin{tabular}{c|cccc}
  & $\bar f_H$& $\bar \nu$  & $\bar g$ & $\bar f$ \\\hline
A&0.150 & 0.211 & 0.188 & 0.150\\
B&0.122 & 0.120 & 0.083 & 0.117\\
D&0.166 & 0.120 & 0.116 & 0.164\\
F&0.161 & 0.211 & 0.208 & 0.166\\
N&0.232 & 0.170 & 0.233 & 0.231\\
S&0.170 & 0.170 & 0.173 & 0.172\\
\end{tabular}
\caption{Comparison of the estimated absolute preference scale using Harker's method~\citep{harker1987incomplete} ($\bar f_H$) and Eq.~\ref{eq:incomplete_ahp1} ($\bar f$).  \label{tb:results}}
\end{table}

\section{Conclusion}

 In this work, we established a novel connection between a popular tool in decision theory; pairwise comparison matrices (PCMs) and non-equilibrium statistical physics. Specifically, we showed that PCMs induce a family of maximum path entropy random walks constrianed to reproduce a non-equilibrium flux.  Notably, only consistent PCMs (incompletely filled or otherwise) induce detailed balanced random walks. Based on these insights, we proposed the entropy production rate $\dot s$ in the induced MERWs as a quantifier of inconsistency. We showed that $\dot s$ satisfies all previously laid out criteria for reasonable consistency indices. We also showed how to use incompletly filled PCMs in Saaty's AHP.

We hope that our work brings together two previously unrelated areas of scientific inquiry namely non-equilibrium Markov processes and pairwise comparison matrices. Notably, recent years have seen a renewed interest in the study of statistical physics of non-equilibrium Markov processes (reviewed by Seifert~\citep{seifert2017stochastic}). For example, many new identities such as various `fluctuation theorems' have been discovered across a wide range of settings. We speculate that the connections established in the current work will allow a greater exchange of ideas between the two previously unrelated fields of inquiry and potentially refine our understanding of consistency in pairwise comparisons.

\section{Acknowledgments} I would like to Thank  Jason Wagoner for stimulating discussions on non-equilibrium flow processes that lead to an investigation into pairwise comparison matrices. I would also like to thank Matteo Brunelli, Luis Vargas, Christian Maes, and Ram Ramanathan for their comments on the manuscript.

%

%

\newpage
\pagebreak

\setcounter{figure}{0}
\makeatletter
\renewcommand{\thefigure}{A\@arabic\c@figure}
\makeatother

\setcounter{table}{0}
\makeatletter
\renewcommand{\thetable}{A\@arabic\c@table}
\makeatother

\setcounter{equation}{0}
\makeatletter
\renewcommand{\theequation}{A\@arabic\c@equation}
\makeatother

\setcounter{section}{0}
\makeatletter
\renewcommand{\thesection}{A\@arabic\c@section}
\makeatother

\section{$\dot s(\gamma)$ satisfies requirements for reasonable consistency indices\label{ap1}}

Recently, Brunelli et al.~\cite{brunelli2015axiomatic,brunelli2017studying} laid out six requirements for any  index $\mathcal I(W)$ that quantifies the inconsistency in a PCM $W$. Here, we show that the entropy production rate $\dot s(\gamma)$ satisfies all of those requirements. They are as follows
\begin{enumerate}
	\item $\mathcal I(W)$ uniquely identifies consistent PCMs. If $\mathcal W$ is a family of cosistent PCMs then $I(W \in \mathcal W) = v^*$ for some $v^* \in { R}$. Conversely,  $\mathcal I(W) = v^* \Rightarrow W\in \mathcal W.$ 
	\item $\mathcal I(PW{\rm Trans}(P)) = \mathcal I(W)$ for any permutation matrix $P$. 
	\item $\mathcal I(W^\gamma) > \mathcal I(W)$ {\it if} $\gamma >1$. Here, $W^\gamma$ denotes element-wise exponentiation.
	\item We start with a consistent PCM $W$. We choose one entry $W_{ab}$ and transform it $W_{ab} \rightarrow W_{ab}^\delta$, $\delta \neq 1$. We also transform $W_{ba} \rightarrow W_{ba}^\delta$. The resultant PCM $W'(\delta)$ is not consistent. We require $\mathcal I(W'(\delta)) >\mathcal I(W'(\epsilon)) > $ if $\delta > \epsilon > 1$. We also require $\mathcal I(W'(\delta)) > \mathcal I(W'(\epsilon)) > $ if $\delta < \epsilon < 1$.
	\item $\mathcal I(W)$ is continuous with respect to entries in $W$.
	\item $\mathcal I(W) = \mathcal I({\rm Trans}(W))$.
\end{enumerate}

We proved that $\dot s(\gamma)$ satisfies requirements  (1), (3), and (6) in the main text. Requirement (2) implies that the entropy production rate in the Markov process is invariant under permutation of vertex labels. $\dot s(\gamma)$ trivially satisfies this requirement as the entropy production rate is the global property of the entire Markov process. $\dot s(\gamma)$ satisfies requirement (5) as well. Since the Perron-Frobenius eigenvalue $\eta(\gamma)$ and the eigenvector $\bar \nu(\gamma)$ are continuous with respect to the elements of $W$~\cite{saaty2012models}, $\dot s(\gamma)$ is also continuous with respect to elements of $W$. 

While we couldn't prove that $\dot s$ satisfies requirement (4), we provide evidence that it is true based on a conjecture that we numerically checked.

Consider two consistent pairwise comparison matrices $W$ and $Q$. We have $W_{ab} > 0 \Rightarrow W_{ab} f_b/f_a$ for some scale $\bar f > 0$ and $Q_{ab} > 0 \Rightarrow Q_{ab} = g_b/g_a$ for some other unrelated scale $\bar g > 0$.  We  assume that the adjacency graph $A$ corresponding to $W$ and $Q$ is identical and is connected. We create two new matrices where a specific entry $W_{ab}$ (and $Q_{ab}$) is changed to $W_{ab} \rightarrow \alpha W_{ab}$ (and $Q_{ab} \rightarrow \alpha Q_{ab}$). We also change the corresponding reciprocal entry $W_{ba}$ (and $Q_{ba}$). Let us call the modified matrices $W(\alpha)$ and $Q(\alpha)$ respectively. Note that $W(\alpha)$ and $Q(\alpha)$ are not consistent if $\alpha \neq 1.$ 

Eq.~\ref{eq:merw_special} of the main text suggests that the induced MERW of a consistent matrix $W$ (and $Q$)  only depends on the properties of the adjacency graph $A$. Hence, the MERW induced by consistent PCMs $W$ and $Q$ are identical. Surprisingly, based on our numerical calculations we observe that the MERWs induced by $W(\alpha)$ and $Q(\alpha)$ are also identical. We conjecture that this is true.

Requirement (4) follows from this conjecture. Consider an incompletely filled but otherwise consistent PCM $W$. Let $A$ denote its adjacency graph. We note that $A$ is also a consistent PCM. As above, let us modify $W_{ab} \rightarrow \alpha W_{ab}$ and $A_{ab} = \alpha A_{ab}$ for some specific entry. We assume that $\alpha \neq 1$.  Let us denote the two modified matrices by $W(\alpha)$ and $A(\alpha)$. Based on our conjecture, the family of MERWs induced by $W(\alpha)$ is identical to the family of MERWs induced by $A(\alpha)~\forall~\alpha$. Consequently $\dot s(W(\alpha)) = \dot s(A(\alpha))$.

Next, let us consider $W(\alpha_1)$ and $W(\alpha_2)$ such that $\alpha_2 > \alpha_1 > 1.$ To prove that the entrop production rate $\dot s$ satisfies requirement (4), we need to show that $\dot s({W(\alpha_2)}) > \dot s({W(\alpha_1)})$. First, we note that $A_{ab}(\alpha_2) = A_{ab}(\alpha_1)^{\tau}$ where $\tau = \log_{\alpha_1} \alpha_2 >1$. Since $\dot s$ satisfies requirement (3), we have $\dot s(A(\alpha_2)) > \dot s(A(\alpha_1)) \Rightarrow \dot s({W(\alpha_2)}) > \dot s({W(\alpha_1)}).$ This proves that $\dot s$ satisfies requirement (4). 

\subsection{Harker's method to fill an incomplete PCM}

\begin{table}[h]
\begin{tabular}{c|ccccccc}
  &A & B & D & F & N & S  \\\hline
A &1 & 1.39 & 0.83 & 0.76 & 0.9 & 0.73\\
B & 0.72 & 1 & 0.74 & 0.87 & 0.50 & 0.77 \\
D & 1.21 & 1.36 & 1 & 0.95 & 0.77 & 0.95 \\
F &  1.32 & 1.15 & 1.05 & 1 & 0.52 & 1.05\\
N & 1.11 & 2.02 & 1.29 & 1.91 & 1& 1.42\\
S & 1.36 & 1.3 & 1.05 & 0.95 & 0.71 & 1
\end{tabular}
\caption{The `filled' PCM using Harker's method. The original PCM is given in Table~\ref{tb:pcm}. \label{tb:pcmH}}
\end{table}

\end{document}